\documentclass[10pt,twocolumn,letterpaper]{article}

\usepackage{cvpr}
\usepackage{times}
\usepackage{epsfig}
\usepackage{graphicx}
\usepackage{amsmath}
\usepackage{amssymb}

\usepackage[pagebackref=false,breaklinks=true,letterpaper=true,colorlinks,bookmarks=false]{hyperref}

\cvprfinalcopy 

\def\cvprPaperID{****} 

\ifcvprfinal\fi
\begin{document}


\title{CAT2000: A Large Scale Fixation Dataset for Boosting Saliency Research}

\author{Ali Borji$^{\dagger}$ \ \ \ \ \ Laurent Itti$^{+}$ \\
$^{\dagger}$Computer Science Department, University of Wisconsin - Milwaukee, USA\\
$^{+}$Department of Computer Science, University of Southern California - Los Angeles, USA\\
{\url{borji@uwm.edu, itti@usc.edu}}
}

\maketitle
\thispagestyle{empty}

\begin{abstract}
\vspace{-10pt}
Saliency modeling has been an active research area in computer vision for about two decades. 
Existing state of the art models perform very well in predicting where people look in natural scenes. 
There is, however, the risk that these models may have been overfitting themselves to available small scale biased datasets, thus trapping the progress in a local minimum. To gain a deeper insight regarding current issues in saliency modeling and to better gauge progress, we recorded eye movements of 120 observers while they freely viewed a large number of naturalistic and artificial images. Our stimuli includes 4000 images; 200 from each of 20 categories covering different types of scenes such as Cartoons, Art, Objects, Low resolution images, Indoor, Outdoor, Jumbled, Random, and Line drawings. We analyze some basic properties of this dataset and compare some successful models. We believe that our dataset opens new challenges for the next generation of saliency models and helps conduct behavioral studies on bottom-up visual attention.
\end{abstract}

\vspace{-10pt}
\section{Introduction \& motivation}
We live in a world where visual data is generated rapidly, continuously and in large volume. The flow of visual data bombarding our retinas needs to be processed efficiently to extract the information that supports our decision making and action selection. To select the important information from the large amount of data received, the nervous system has to intelligently filter its inputs. The same problem is faced by many computer vision systems especially if they have to function in real time. 

To understand how humans select information and perceive scenes, researchers usually record eye movements of people while they freely watch images~\cite{IT98,borji2013state,borji2013quantitative}. Recent saliency models perform very well, almost close to human inter-observer model, in predicting fixations. However, these models have been evaluated over biased fixation datasets. Existing datasets, since it is expensive to collect fixation data, have small number of scenes shown to few observers. Further, stimulus variety is limited in existing datasets and often objects appear at the center of scenes (center-bias). To tackle these shortcomings, some researchers resort to webcams and clicks through Amazon Mechanical Turk but it is difficult to control the quality of the collected data in this manner (e.g., eye tracking accuracy and calibration, observer distance and field of view, mood, age, intelligence, concentration, etc.). Thus challenges regarding dataset bias need to be properly addressed in the saliency modeling  similar to other areas in computer vision~\cite{torralba2011unbiased}. To this end, we systematically collect a large scale fixation dataset over several categories of images.

\begin{figure*}[htbp!]
\centering	
\includegraphics[scale=0.17]{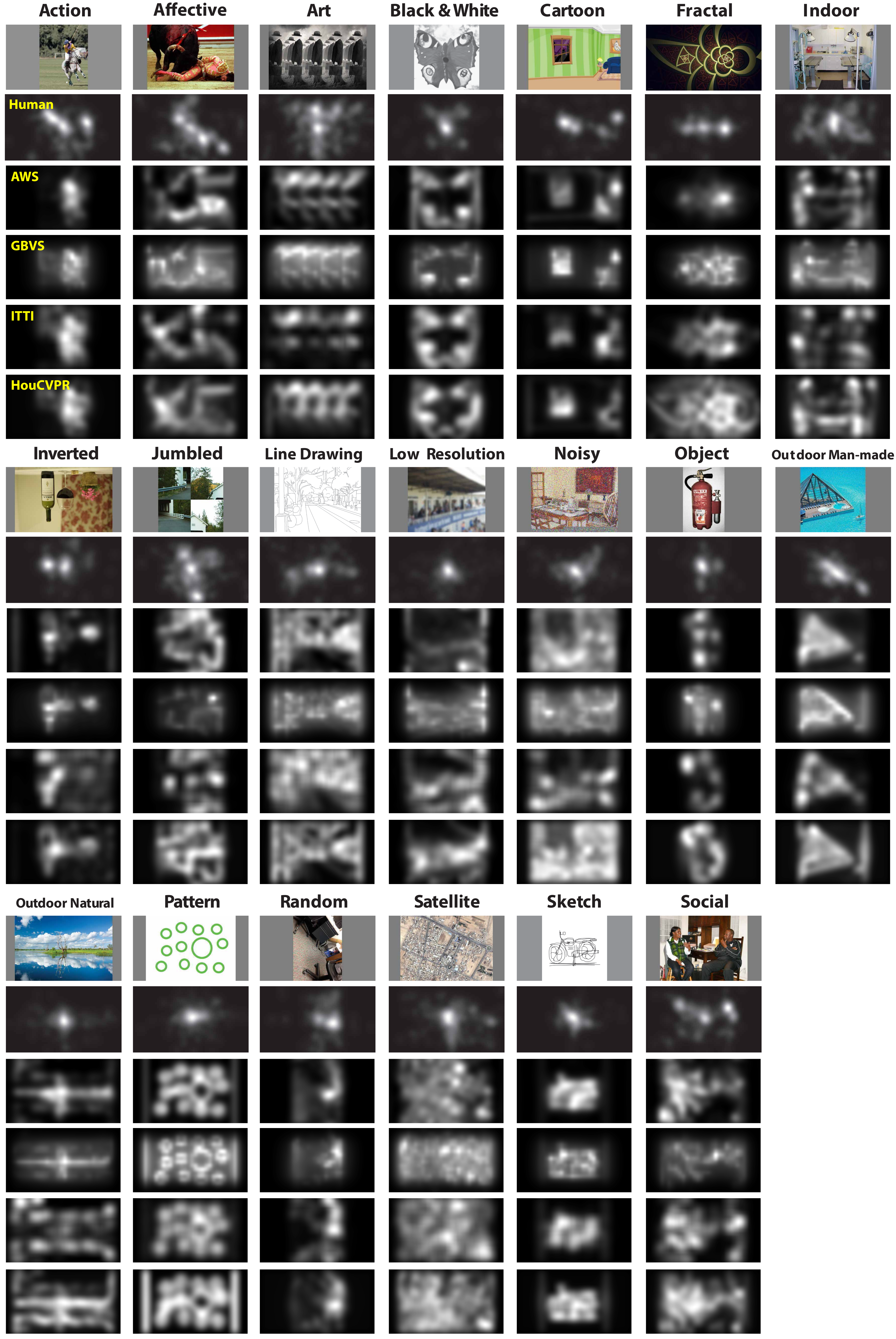}
\caption{Sample images from our dataset along with fixations and saliency maps (one sample per category).}
\label{fig:salMaps}
\end{figure*}

\section{CAT2000 dataset}
\subsection{Stimuli}

We have scenes from 20 categories including: 1) Action, 2) Affective, 3) Art, 4) Black \& White, 5) Cartoon, 6) Fractal, 7) Indoor, 8) Inverted, 9) Jumbled, 10) Line drawings, 11) Low resolution, 12) Noisy, 13) Object, 14) Outdoor man made, 15) Outdoor natural, 16) Pattern, 17) Random, 18) Satellite, 19) Sketch, and 20) Social. Images have resolution of $1920 \times 1080$ pixels. Fig.~\ref{fig:salMaps} shows an example from each category along with human fixations. Some of our categories elicit bottom-up (BU) attention cues strongly (e.g., Pattern) while others elicit top-down (TD) factors significantly (e.g., Social). Thus different categories are suitable for studying different aspects of attentional behavior. To collect images, we used Bing and Google search engines to retrieve images using several relevant key terms as well as some computer vision datasets

\textbf{Objects} were 200 categories of the Caltech256 dataset. We randomly chose one sample from each category. 
\textbf{Outdoor man made} category basically contains images of structures built in open space (e.g., building, road, bridge, ferris wheel). 
\textbf{Outdoor natural} include scenes from the nature (e.g., lilies, mountains, forest, animals). 
\textbf{Pattern} contains 200 psychological patterns which have often been used for evaluation of BU saliency models mainly in behavioral studies including pop-out, conjunction, search asymmetry, etc. 
We included a set of texture defects as well. \textbf{Random:} It is well known that humans look at the center of the screen mainly due to photographer bias and partly due to viewing strategy in desktop eye tracking setups. To handle this hurdle, we captured some images from random viewpoints using a cell-phone camera with closed eyes and random orientations. \textbf{Satellite:} We used MS Bing maps and saved images from different random geographical locations. Note that altitude is also chosen randomly making some images aerial and some satellite view. \textbf{Sketch} category contains sketches of 200 objects (similar to the object category) taken from the EITZ dataset~\cite{eitz2012hdhso}. These images, similar to line drawings, contain no color and texture. The \textbf{Social} category contains pictures of people having social interaction. Our aim here was to study the high-level semantic attentional cues (e.g., gaze direction). For \textbf{Action} category, we used some images from the Stanford action dataset~\cite{yao2011human}. For \textbf{Indoor and Outdoor man made}, we used 15 scenes and SUN datasets. For \textbf{Line drawing}, we used the Lotus Hill dataset with 6 categories. To build the \textbf{Jumbled} category, we randomly divided each scene into $n \times m$ partitions (n and m chosen randomly from $\{2,3,4,5\}$). Partitions were then randomly shuffled. \textbf{Affective} category contains emotional scenes with mild versions of scary, disgust, joy, happiness, sadness, anger, violence, etc. \textbf{Art} scenes contain artworks and paintings from different styles, computer-generated arts, man-made artifacts/designs, etc. \textbf{Cartoons} are mainly computer generated scenes with non-natural renderings and often contain simple entities. \textbf{Fractal} category contains synthetic images from different types of fractals.
\textbf{Low resolution scenes} were taken from Judd et al.~\cite{judd2011fixations} (Gaussian blurred). \textbf{Noisy} images were generated by adding Gaussian, Speckle, and Salt \& Pepper noises. \textbf{Black \& White} category contains gray scale images.


\subsection{Observers}

We had 120 observers (40 male, 80 female) in total. Mean observer age was 20.15 (min = 18, max = 27m std 1.65, median 20). Observers were undergraduates at USC from different majors and from mixed ethnicities. The experimental methods were approved by USC's Institutional Review Board (IRB). Observers had normal or corrected-to-normal vision and received course credit for participation. They were naive to the purpose of the experiment and had not previously seen the stimuli. 

\begin{figure*}[t]
\centering	
\vspace{-10pt}
\includegraphics[scale=0.65]{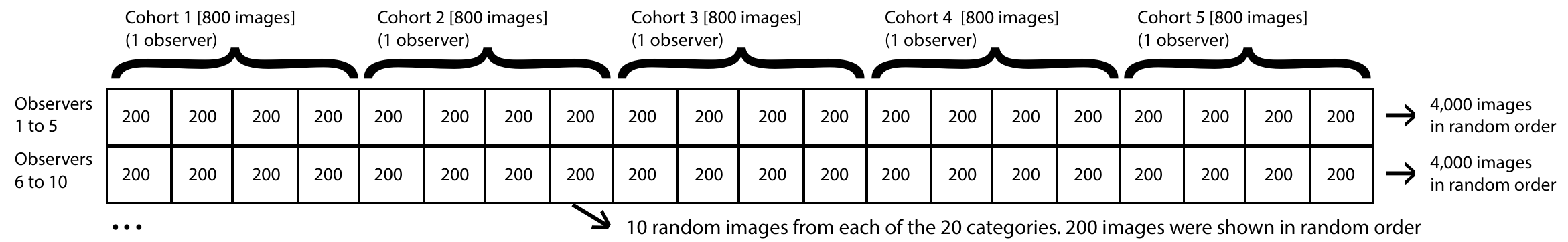}
\caption{Distribution of images across observers.}
\label{fig:setting}
\vspace{-10pt}
\end{figure*}

Fig.~\ref{fig:setting} shows the assignment of observers to stimuli and the way we conducted the experiment. Since it was not possible to show all images to an observer, we partitioned 4,000 images into five cohorts each of size 800. Each cohort was further divided into 4 sections of 200 which was shown to an observer (one section per session). Each section last about 25 minutes followed by 5 minutes rest. The eye tracker was re-calibrated at the beginning of each recording session (i.e., 200 images). Each observer viewed all images in a cohort (in 4 sessions). All 4,000 images were randomly shuffled with the constraint that each section must include 10 images from each category (i.e., 10 $\times$ 20). Each image was viewed by 24 different observers. We had 24 passes (showing all 4,000 images) over all data, each pass by 5 observers (i.e., 120 observers = 5 $\times$ 24 passes).

\subsection{Eye tracking procedure}
Each trial began with a fixation
cross at the center of a blank screen, which observers were instructed to fixate. Following the fixation
cross, a target image was shown for 5 seconds followed by 3 seconds gray screen. Observers were instructed to "look around the image" with no restrictions.

Observers sat 106 cm away from a 42 inch LCD monitor screen such that scenes subtended approximately $45.5^{\circ} \times 31^{\circ}$ of visual angle (degree of visual angle or dva about 38 pixels). A chin rest was used to stabilize head movements. Stimuli were presented at 60Hz at a
resolution of $1920 \times 1080$ pixels (with added gray margins while preserving the aspect ratio; see Fig.~\ref{fig:salMaps}). Eye movements were recorded via a non-invasive infrared Eyelink (SR Research, Osgoode, ON, Canada) eye-tracking device at a sample rate of 1000 Hz (spatial resolution less than 0.5$^{\circ}$). The eye tracker was calibrated using 5 points calibration at the beginning of each recording session. Saccades were classified as events where eye velocity was greater than $35^{\circ}$/s and eye acceleration exceeded $9500^{\circ}$/$s^2$ as recommended by the manufacturer for the Eyelink-1000 device. 

\section{Dataset statistics \& model comparison}

First, we analyze some basic properties of our dataset.
In total, we have 24,148,768 saccades over 240 hours of viewing time.
We find that some categories are more center-biased (e.g., Action, Affective, Art, Black \& White, Fractal, Line Drawing, Low Resolution, Noisy, Object, Pattern, and Sketch) compared to others (e.g., Cartoon, Indoor, Inverted, Jumbled, Outdoor Man-made, Outdoor Natural, Random, Satellite, and Social). The reason for high center-bias 
over some categories might be photographer bias (e.g., Action, Objects, Sketch) or less interestingness (e.g., Noisy, Low Resolution, Line Drawings). The reason for less center-bias over some categories might be the spread of content over the entire scene (e.g., Random, Outdoor Man-made, Jumbled, Satellite, and Social).  

Median number of saccades per image is around 20 over all subjects and categories for 5 seconds viewing. The variance is about 6 saccades. The mean number of saccades for some categories such as Low-Resolution, Noisy, Sketch, and Pattern is lower than others (such as Social, Jumbled, Affective, and Cartoon).

To analyze the degree of observer consistency over categories, we measure the inter-observer (IO) agreement on each image. For each image, one of the 24 observers is set aside. The smoothed map of fixations of all other observers is then applied for the predicting fixations of the remaining observer. The prediction power is measured using the Normalized Scanpath Saliency (NSS) score which is the average map activations at fixations in the normalized map (zero mean, unit standard deviation). Categories with high IO score (thus high observer consistency) include Sketch, Low Resolution, Affective, and Black \& White. Categories with low IO score include Jumbled, Satellite, Indoor, Cartoon, and Inverted. Note that categories with high center-bias usually result in higher IO consistency.

\begin{figure}[htbp!]
\centering	
\vspace{-5pt}
\includegraphics[scale=0.5]{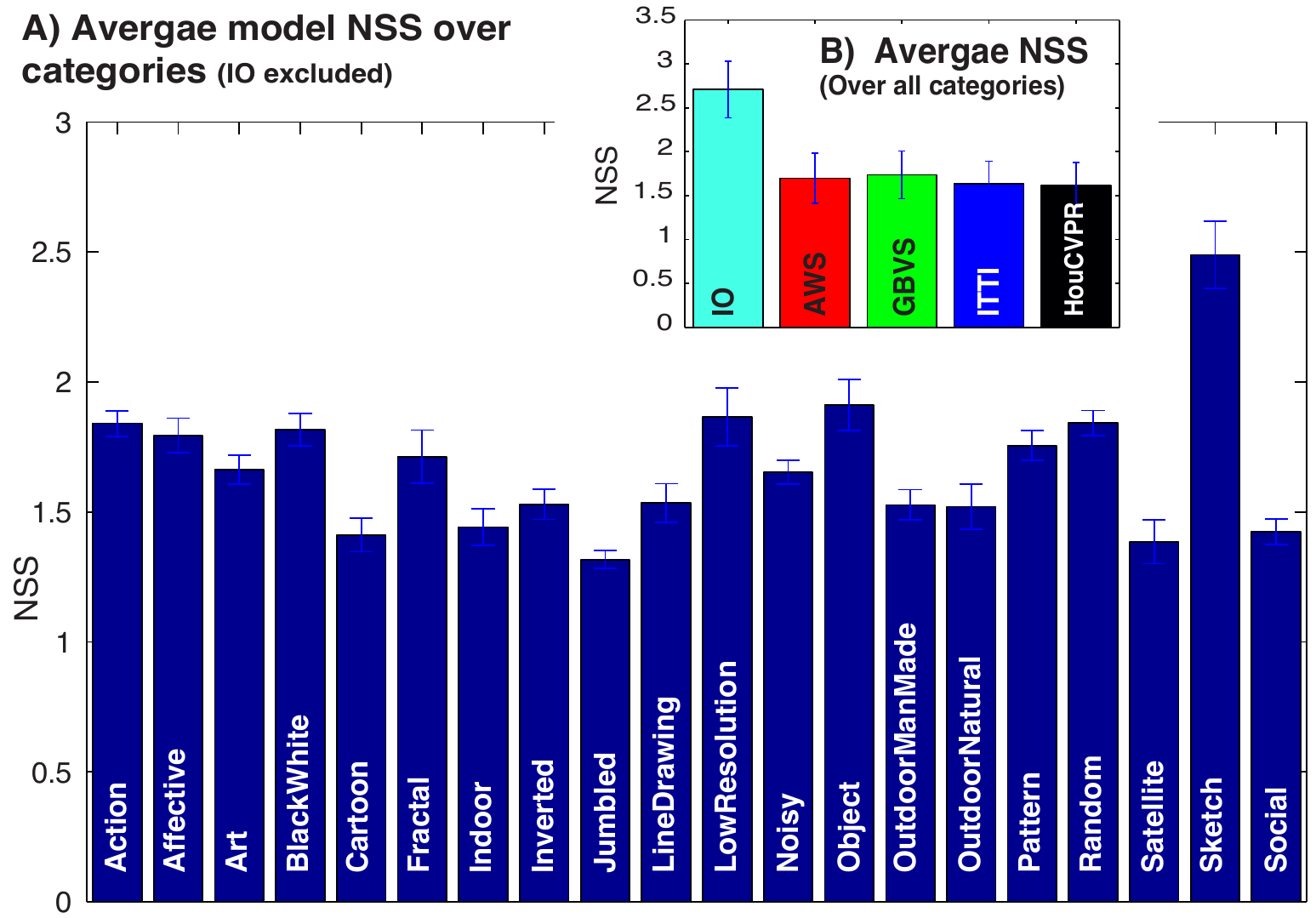}
\caption{Model performance across categories.}
\label{fig:GrandAvg}
\vspace{-15pt}
\end{figure}

Next, we evaluate performance of 4 popular saliency models including ITTI~\cite{IT98}, HouCVPR
~\cite{SR08}, GBVS~\cite{GB06}, and AWS~\cite{GarciaDiazJOV} over our dataset. Fig.~\ref{fig:GrandAvg} shows the average NSS score for each model over 20 categories. To our surprise, models did very well over the Sketch category (1st rank) but poorly over the line drawings. The reason can be because line drawings have content across the image while sketches contain objects at the center. This makes models generate more activation at the image center which matches better with focused fixations at the center of the object. Some other difficult categories include Social, Satellite, Jumbled, and Cartoon categories. There are different reasons for different categories. For example, some top-down cues might affect fixations while models don't account for them (e.g., gaze direction over social scenes). Satellite images might have been boring for observers causing more center-bias while models generate activation everywhere. Some models are affected by block borders over Jumbled images while humans discard them. Humans and models might be biased toward viewing upright images causing performance degradation over inverted images. Note that scores are averaged over all 4 models here. Models did well over Object, Low Resolution, Random, and Action categories. Fig.~\ref{fig:GrandAvg} inset shows NSS scores across all categories for each model. We find that models perform about the same and all score significantly below the IO model.

\section{Discussion \& conclusion}
We introduced a large scale eye movement dataset containing 4000 images from a variety of categories. Here, we scratched the surface of this dataset. In addition to benchmarking purposes, our dataset can be used to conduct behavioral studies and to investigate semantic cues that may guide gaze in free viewing of natural scenes.

To make this dataset available to public for model benchmarking\footnote{Available at: http://saliency.mit.edu/results\_cat2000.html}, we have divided it into two sets of images: train and test. Train images (100 from each category) and fixations of 18 observers are shared but 6 observers are held-out. Test images are available but fixations of all 24 observers are held out. In this way, researchers can train their models to predict fixations of new observers on the same images (seen by others) or on totally unseen images.

\vspace{5pt}
{{\small
\noindent \textbf{Acknowledgments:}
We wish to thank Deborah Lee, Nitika Jawaheri, Jiawei Wang and Noa Shemi, undergraduates at University of Southern California, for their help on collecting this dataset.
}

{\footnotesize
\bibliographystyle{ieee}
\vspace{-6pt}
\bibliography{egbib}

}

\end{document}